\documentclass[lettersize,journal]{IEEEtran}
\usepackage{amsmath,amsfonts}
\usepackage{algorithmic}
\usepackage{algorithm}
\usepackage{array}
\usepackage[caption=false,font=normalsize,labelfont=sf,textfont=sf]{subfig}
\usepackage{textcomp}
\usepackage{stfloats}
\usepackage{url}
\usepackage{colortbl}
\usepackage{verbatim}
\usepackage{graphicx}
\usepackage{cite}
\usepackage{amsmath,amssymb,amsthm,amsfonts,latexsym,bbm,xspace,graphicx,float,mathtools,
verbatim,xcolor} 
\usepackage{amsmath}
\newtheorem{theorem}{Theorem}

\newtheorem{lemma}[theorem]{Lemma}
\newtheorem{prop}[theorem]{Proposition}
\newtheorem{cor}{Corollary}
\newtheorem{assumption}{Assumption}

\newtheorem{defn}{Definition}

\begin{document}

\title{Walk for Learning:\\
A Random Walk Approach for Federated Learning from Heterogeneous Data}

\author{

  \IEEEauthorblockN{Ghadir Ayache\IEEEauthorrefmark{1}, Venkat Dassari\IEEEauthorrefmark{2},  Salim El Rouayheb\IEEEauthorrefmark{1}}
  \\
    \IEEEauthorblockA{\IEEEauthorrefmark{1}Electrical and Computer Engineering Department, Rutgers University
  }
  \\
    \IEEEauthorblockA{\IEEEauthorrefmark{2}US Army Research Lab
 }


}



\maketitle

\begin{abstract}
We consider the problem of a Parameter Server (PS) that wishes to learn a model  that fits data distributed on the nodes of a graph.
We focus on Federated Learning (FL) as a canonical application. One of the main challenges of FL is the communication bottleneck between the nodes and the parameter server. A popular solution in the literature is to allow each node to do several local updates on the model in each iteration before sending it back to the PS.  While this mitigates the communication bottleneck, the statistical heterogeneity of the data owned by the different   nodes has proven to delay   convergence and  bias the model. 

In this work, we study random walk (RW)  learning algorithms for tackling the communication and data heterogeneity problems. The main idea is to leverage   available direct connections among the nodes themselves,  which are typically ``cheaper" than the communication to the PS. 
In a random walk, the model is thought of as a ``baton" that is passed from a node to one of its neighbors after being updated in each iteration.   

The challenge in designing the RW is the data heterogeneity and the uncertainty about the data distributions.  
Ideally, we would want to visit more often nodes that hold more informative  data.  We cast this problem as   a sleeping multi-armed bandit (MAB)
to design near-optimal node sampling strategy that achieves a variance reduced gradient estimates and approaches sub-linearly the optimal sampling strategy. 
Based on this framework, we present an adaptive random walk learning algorithm. We provide theoretical guarantees on its convergence. 
Our numerical results  validate our theoretical findings and  show that our algorithm outperforms existing random walk algorithms.  
\end{abstract}


\section{Introduction}
\label{submission}
\subsection{Overview and Motivation}

\begin{figure}
    \centering
    \includegraphics[scale=0.18]{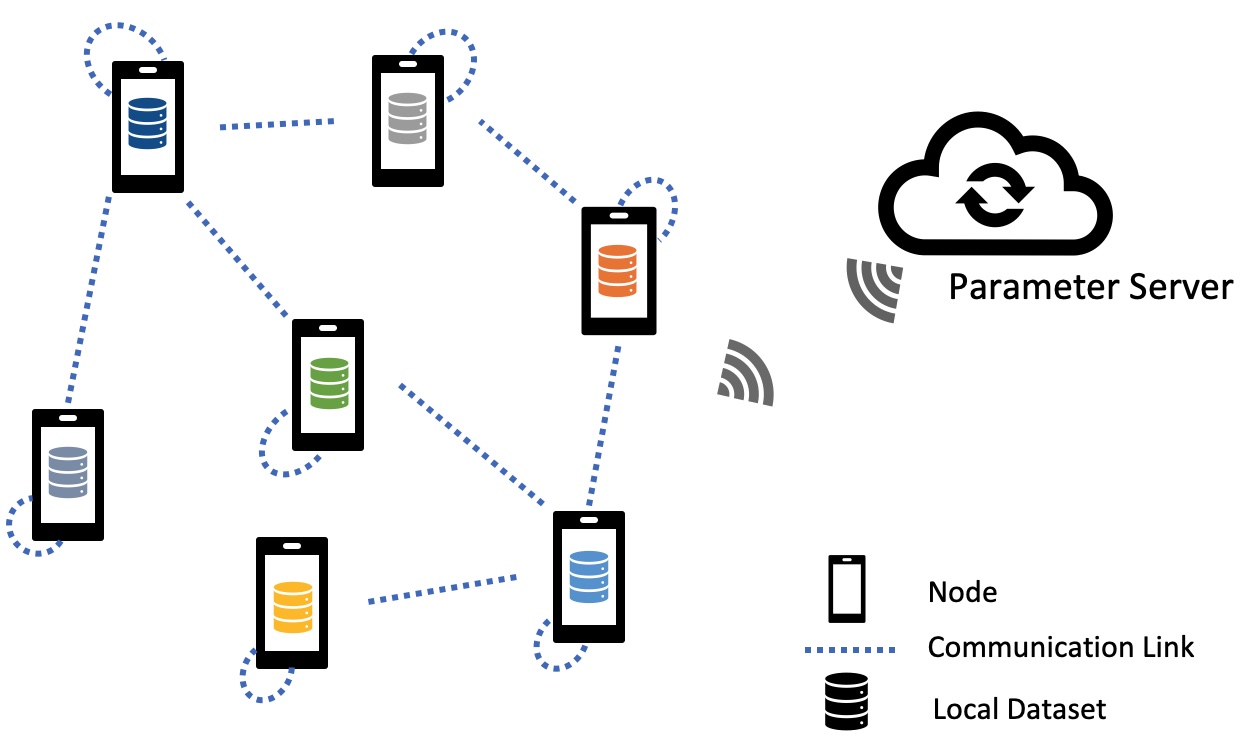}
    \caption{{Distributed Learning on Graph through local computations at the graph's nodes and through local communication between the connected nodes.}}
    \label{FL}
\end{figure}

\IEEEPARstart{D}{istributed} Machine Learning has proven to be an important framework for training machine learning models without moving the available data  from its local devices, which ensures privacy and scalability. Federated Learning (FL) has risen to be one of the main applications \cite{FL,kairouz2019advances,Li2020FederatedLC} 
that has been attracting significant research attention and has been deployed in real-world  systems with millions of uses \cite{kairouz2019advances}.
Other applications include  learning in  IoT networks \cite{YangIoT}, smart cities and healthcare \cite{Brisimi2018FederatedLO,jiang2020federated}. 
To see how a typical learning algorithm works in this setting, consider the FL    setting    in Fig~\ref{FL}. There is as Parameter Server (PS) (typically sitting in the cloud) and a number of nodes (phones, IoT-devices, smart sensors, etc.) each having its own local data. The PS wishes to learn a global model on all the data  without moving the data away from its original owner. The algorithm would work in a batch SGD fashion. In each iteration, the PS samples a batch of nodes and sends the current model to it. Each node in this batch will update the model based on its local data and sends back its updated model to the PS. The PS then   aggregates all the received models and starts over again. 

\vspace{1em}

\noindent {{\bf Locality vs.\ Heterogenity.}} One of the main bottlenecks here is the communication with the parameter server (PS) needed in each iteration to aggregate the updates sent by the nodes   and to coordinate the learning process. A popular solution is  to reduce the communication cost with the PS is to let each node perform  several local model updates on its data before reporting back to the PS \cite{mcmahan17a}.  However, the local computations may   induce local biases to the model and slow the convergence of the learning algorithm. This is due to the   data  that is heterogeneous across the different nodes, which imposes an inconsistency between the local and the global objectives \cite{zhao2018federated,wang2020tackling}.







 
\vspace{1em}

\noindent{\bf  Random Walks.}
\begin{figure}
    \centering
    \includegraphics[scale=0.18]{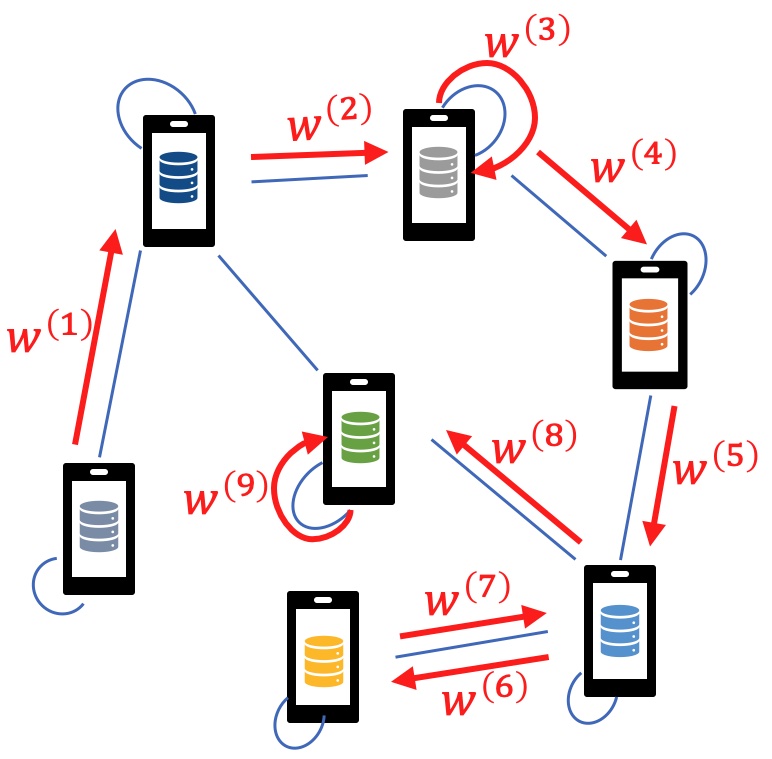}
    \caption{Random Walk (RW) on the graph: the updated model $w^{(k)}$ is transmitted at time $k$. The self-loops at $k =3$ and $k=9$, indicate that the RW algorithm decided to make a second update at the same node. The model is reported back to the PS on a regular basis.}
    \label{RW}
\end{figure} We propose Random Walks (RW) as a way to simultaneously achieve two seemingly opposing goals: extending the benefits of locality and mitigating the drawbacks of data heterogeneity.
The idea being that instead of restricting local computations to the  node itself, they can be extended to its neighboring nodes. This is achieved by leveraging existing local connections among the nodes themselves, which are typically cheaper than communication to the PS.   
We represent the local connections by a graph structure where each node is connected to a subset of other neighboring nodes. Thus, each node can exchange  information with its neighbors  and, through these local communications, information can propagate  through the whole network. This setting can arise in mobile and edge networks,  IoT applications and ad-hoc networks, to name a few.
 

In our proposed framework,  the learning algorithm will run  as a random walk where in each iteration the model gets updated at one of the nodes and then passed to one of its neighbors   to take over the next update, as shown in Figure~\ref{RW}. The model is then passed to the PS on a regular basis and/or depending on the network resources to mitigate the communication cost to the PS.

Random walk learning algorithms have  been well studied in  the literature on optimization \cite{Nedic2009DistributedSM, Johansson2009ARI}, wireless networks \cite{zhang2018compressive} and signal processing \cite{Mao2020WalkmanAC}.   What distinguishes this work is that it tackles the problem of designing random walk learning algorithms in the presence of arbitrary data heterogeneity across the nodes. Our main contribution is a      random walk algorithm that, along with updating the model, it learns and   adapts to the different nodes' distributions by carefully combining   exploration and exploitation. Our main tool is the theory of Multi-Armed Bandit (MAB). 






\vspace{1em}

\noindent{\bf Random Walk via Multi-Armed Bandit.}{ Typically, the distributions of the local data at each node are not known a priori.
Therefore, we want to    
  devise a random walk strategy that overcomes data heterogeneity by learning about the local data distributions along the way. The goal is to minimize the variance of the global objective gradient estimates computed locally by adjusting the nodes' sampling strategy.  More specifically, at each iteration $k$, one has to design the probabilities with which the next node in the RW is chosen among the neighboring nodes. Note that these probabilities will depend on $k$ to adapt to the information learned so far, leading to a time-varying RW. Therefore, the random walk will start with an exploration phase before gradually transitioning into an exploitation phase, once more robust  estimates about the nodes' data are obtained.
  
  We design the RW by casting our problem as Mulit-Armed bandit. The multi-armed bandit (MAB) is a learning framework to decide optimally under uncertainty \cite{auer2002finite,auer2002nonstochastic,bubeck2012regret}. It features $N$ arms with unknown random costs (negative rewards). At each iteration, one pulls.    The problem is to decide on which arm to pull each time   so that the accumulated cost, called {\em regret}, after playing   $T$ times is minimized. Our 
  solution is an algorithm that explores the different arms and, in parallel, it exploits the collected information so far, where it observes the outcome of playing an arm and uses it to tune its expected reward estimate to adjust future selections \cite{auer2002finite, auer2002nonstochastic}. 
In our distributed learning setting, we model the node selection in the RW as arm pulling in the MAB framework. At each iteration, the random walk picks a node in graph to activate for the next update, observes the update, and receives the local gradient as a cost.


Under this analogy, the performance of the learning algorithm is measured by the regret,   which is the difference between the cost of a random walk with optimal transition matrix  when all the distributions are known, and the cost of the nodes visited by the algorithm. 


}

\subsection{Contribution}





In this paragraph, we summarize our contribution as follows:
\begin{itemize}
    \item In this work, we propose a distributed learning algorithm to learn a model on the distributed data over the nodes in a graph. Our algorithm selects the nodes to update the model by an adaptive random walk on the network to address the statistical heterogeneity of  distributed data.
    
    \item We model the random walk transition design as a sleeping multi-armed bandit problem to compete with the optimal transition probabilities that mitigate the high variance in the local gradients estimates.
    
    
    
    \item  We provide the theoretical guarantee on the rate of convergence of our proposed algorithm approaching a rate $\mathcal{O}(1/\sqrt{T})$. The rate depends on the graph spectral property and the minimal gradient variance.
    
    \item  Finally, we simulate our algorithm on real and synthetic data, for different graph settings and heterogeneity levels and show that it outperforms existing baseline random walk designs.
\end{itemize}


    
    
    
    
    

\subsection{Prior Work}



\noindent {\bf  Random Walk Learning.} 
Several works have studied random walk learning algorithms   focusing on  the convergence under different sets of assumptions. The works of \cite{Johansson2007ASP,Johansson2009ARI,Ram2009AsynchronousGA}   established theoretical convergence guarantees for uniform random walks for different convex problem settings and using first-order methods. Later work \cite{Wai2017CurvatureaidedIA}  employs more advanced stochastic updates based on  gradient tracking technique that uses Hessian
information to accelerate convergence. The work of \cite{MCGD} proposed to speed-up the convergence by using non-reversible random walks. In \cite{Mao2020WalkmanAC}, the authors studied the convergence of   random walks learning   for the alternating direction method of multipliers (ADMM). In \cite{ayache2020}, the paper proposes to improve the convergence guarantees by designing a weighted random walk   that accounts for the importance of the local data to speed up convergence.
An asymptotic fundamental bound on the convergence rate of these algorithms was proven by \cite{ Duchi2011ErgodicMD} and it approaches $O(1/\sqrt{k})$ under convexity and bounded-gradient assumptions. 
From our MAB perspective,  a common aspect of these algorithms is that they are purely   exploitative.  They require  a priori information about the local data (e.g., gradient-Lipschitz constants, bounds on the gradients) to design a non-adaptive (time-invariant) random walk. 



Random Walk algorithms belong to a more general class of decentralized learning algorithms where no central entity, such as a PS,  is involved to handle the learning process. Gossip algorithms are another class of decentralized algorithms that are not based on random walks   (see for e.g.\ \cite{Nedic2009DistributedSM, Shi2015EXTRAAE, Chen2012FastDF,Jakoveti2014FastDG}).  In   (synchronous) gossip algorithms, at each round,  each node  updates and exchanges its {\em local model}  with its neighboring nodes. Hence, in each iteration, all the nodes and all the links in the graph are activated. The goal of a Gossip algorithm  is to ensure that all nodes, and not just the PS, learn the global model and assume convergence once a consensus is reached. Hence, it is less efficient in terms of computations and communication costs   \cite{ayache2020}.


%



\vspace{1em}

\noindent {\bf  Data Heterogeneity.} 
Recently, there has been lots of work addressing the problem of data heterogeneity especially in the FL literature.
The data across the nodes is typically not generated in an iid fashion.
A local dataset  tends  to be more personalized and biased towards its specific owner's profile. Therefore, multiple local updates on the global model can drift  the global objective optimization towards its local one. This may slow down  the convergence   and can lead to converging to a suboptimal model \cite{li2018federated, wang2020tackling}. 
  Several measures have been proposed in the literature to quantify   statistical heterogeneity, which refers to  this local vs.\ global objectives' inconsistency in the distributed data.
The focus has been on quantifying the gap between the local update direction  and the global one \cite{zhao2018federated,li2018federated,haddadpour2019convergence,karimireddy2020scaffold,khaled2019first}. The proposed solutions vary between controlling the update direction \cite{karimireddy2020scaffold} or the learning objective \cite{li2018federated}. 

\vspace{1em}

\noindent {\bf  Multi-Armed Bandit Sampler. } The multi-armed bandit (MAB) problem aims to devise optimal   sampling strategies by balancing together exploration and exploitation  \cite{auer2002finite,auer2002nonstochastic,Kleinberg2010RegretBF,bubeck2012regret}.
Results from MAB have been used in problems related to  
  standard  SGD training in a non-distributed settings \cite{Namkoong2017AdaptiveSP,Salehi2017StochasticOW,Borsos2018OnlineVR,Zhang2018AdaptiveOL, Hanchi2020AdaptiveIS}. The idea there is to use an  MAB sampler to select more often the data points that can better guide the learning algorithm.
  
  In the classical MAB setting there is no constraint on which node to sample (visit) at a given time (which arm to pull in the MAB language). However, in our case, we are restricted by the graph topology, so only neighboring nodes can be visited. 
 To account for this constraint, we cast our problem as     Sleeping Multi-Armed Bandit  in which  nodes that are not neighbors of the current nodes are assumed to be sleeping (not available) at the time of the sampling. The Sleeping MAB literature has studied various assumptions on sleeping reliance: independent availabilities, general availabilities, and adversarial availabilities. The lower bound on the regret is known to be $\Omega(\sqrt{NT})$ if we consider stochastic independent availabilities. For a harder sleeping-MAB setting with adversarial availabilities, the lower bound is $\Omega(N\sqrt{T})$ for $N$ being the total number of nodes and $T$ being the total number of rounds \cite{Kanade2009SleepingEA,Kleinberg2010RegretBF,Saha2020ImprovedSB}. In our work, we model the RW design problem as dependent availabilities Sleeping MAB learning algorithm \cite{Saha2020ImprovedSB}. For the proof technicality, we use a harder upper bound on the performance that assume oblivious adversarial availibilities.

A   related line of work is the work on importance sampling which can be thought of as a pure exploitation scheme  with no exploration. The literature has studied different aspects of importance sampling using prior information on the local datasets (e.g.,\cite{Zhao2014StochasticOW, Alain2015VarianceRI,Stich2017SafeAI, Papa2015AdaptiveSF}). For instance, in \cite{Needell}, the paper proposes to sample proportionally to the smoothness bounds of the local objectives. While the scheme in  \cite{Zhao2014StochasticOW} suggests to select the data points based on the bounds of the gradients of the local objectives. 
\subsection{Organization}

The rest of the paper is organized as follows.  We present the problem setup  in  Section~\ref{sectionII}.  In Section~\ref{sectionRW}, we present the detailed random walk learning algorithm. In Section~\ref{sectionNodeSampling}, we provide the optimal sampling scheme and its theoretical motivation. In Section~\ref{SectionSleepingSampling}, we outline the analogy between the RW design problem and the sleeping MAB problem. In Section~\ref{sectionResults}, we  present the MAB RW learning algorithm and the main theorem on its convergence. Moreover, we provide the technical definitions and assumptions used into the main theorem proof in Section~\ref{sectionproof} .   Finally, we provide   numerical results on the convergence of our proposed algorithm   in Section~\ref{sectionVI}. 
The full proofs of the technical results are deferred to the appendices.

\section{Setup} 
\label{sectionII}

\subsection{Network Model.}
We represent a network of $N$ nodes by 
 an undirected graph $\mathcal{G}(V,\,E)$ with $V = \{1,\,...,\,N\}$ being the set of nodes and $E$ being the set of edges such that  $E = \{ (i,j) \in V \times V, \text{ if } i  \text{ is connected to }  j\}$. Since the graph is undirected, then $\forall (i,j) \in E, \text{ we have } (j,i) \in E$. Any two connected nodes $i$ and $j$ are called neighbor nodes and we denote it by $i\sim j$. Moreover, we assume that all the nodes have  self-loops, thus, $\forall i,\, (i,i) \in E$.

\subsection{Data Model.}
We assume that  every node $i$   owns a dataset $\mathcal{D}_i$ of size $n$  such that ${\mathcal{D}}_i=\left\{ \xi_{i,j}:=(x_{i,j},\,y_{i,j}) \in \mathbb{R}^d \times \mathbb{R} \quad
 \text{for} \quad  j \in \left[n\right]\right\}$, which is sampled from an unknown local distribution $\Pi_i$.


  

\subsection{Learning Objective.} Our goal is to minimize a global objective function $F(w)$ where $w \in \mathcal{W} \subset \mathbb{R}^d$, $\mathcal{W}$ being the feasible set   assumed to be   closed and bounded. The objective function $F(.)$  represents the empirical mean of local losses on the distributed data over the  graph of $N$ nodes. Therefore, we are looking to solve the following problem: 
\begin{align}
\label{eq:1}
    \underset{w\in\mathcal{W}}{\min} \left \{ F(w)  \coloneqq \frac{1}{N} \sum_{i=1}^{N} F_i\left(w\right) \right\},
\end{align}
where the function $F_i$ is the local objective at the node $i$ and it is defined as 
 \begin{equation}F_{i}\left(w\right)=\mathbb{E}_{\xi_{i}}\left[F_{i}\left(w;\xi_{i}\right)\right]
  \text{ for  } \, \xi_{i} \sim  \Pi_i. \end{equation}
 The optimal model is denoted by $w^*$ and defined as follows:
\begin{align*}
       w^* = \underset{w\in\mathcal{W}}{\arg\min \,}  F(w).
\end{align*}

   \subsection{Data Heterogeneity.}The data distributions  across the nodes of the network are assumed to be arbitrary. Therefore, when a node performs a local update on the global model it may  bias  it  by its   local dataset that may not be a good representative of  the global learning objective. Multiple definitions have been recently proposed   to quantify the degree of local  heterogeneity in distributed systems \cite{li2018federated,khaled2019first,wang2020tackling,haddadpour2019convergence}. These   definitions focus on the variance of the  local gradients with respect to the global gradient at a given model $w$. In our work, we adopt the definition used by \cite{karimireddy2020scaffold} as   stated below.



\begin{defn}[Data Heterogeneity] The local objectives $F_i$s are $(\alpha,\sigma)$-locally dissimilar at $w$ if 
\begin{align*}
    \mathbb{E}_{i}\left[\left\Vert {\nabla}F_{i}\left(w\right)\right\Vert _{2}^{2}\right]\leq\alpha^{2}+\sigma^{2}\left\Vert \nabla F\left(w\right)\right\Vert _{2}^{2}, 
\end{align*}
for $\alpha \geq 0 $, $\sigma \geq 1$, $ \mathbb{E}_{i}$ is the expectation over the nodes and ${\nabla}F_{i}\left(w\right)$ is the gradient at node $i$. For $\alpha=0$ and $\sigma=1$, we restore the homogeneous case.

\label{def:1}
\end{defn}
 This definition is a generalization of other definitions  \cite{li2018federated,khaled2019first}. 

 

 \subsection{Model Update.}
 We focus  in   our analysis on first-order methods using stochastic gradient descent\footnote{Our work is applicable to any iterative algorithm that uses an unbiased descent update \cite{Bottou2007TheTO,Defazio2014SAGAAF,Xiao2014APS}.}.  Given the limited compute power of the nodes,
 
 Thus, the model update at round $k$ will be as follows
 \begin{align}
\label{eq:2}
 w^{(k+1)}= {\bf{\Pi}}_{\mathcal{W}}\left(w^{(k)}-\gamma^{(k)}
 \frac{1}{p^{(k)}(i^{(k)})}\hat{\nabla}  F_{i^{(k)}}\left(w^{(k)}\right)\right),
 \end{align}  where $\gamma^{(k)}$ is the step size, $\hat{\nabla} F_{i^{(k)}}\left(w^{(k)}\right)$ is an unbiased estimate of the local gradient at node $i^{(k)}$ computed on a uniformly sampled data point from $\mathcal{D}_{i^{(k)}}$, ${p^{(k)}(i^{(k)})}$ is the probability of picking node $i^{(k)}$ at round $k$, and ${\bf{\Pi}}_{\mathcal{W}}$ is the projection operator onto the feasible set ${\mathcal{W}}$. 
  For our convergence analysis, we will need   the following technical assumptions.
 
 \begin{assumption}
 \label{ass1}
 For every node $i \in [N]$, the local loss function $F_i(.):\mathcal{W}\rightarrow \mathbb{R}^d $ is differentiable and  convex function on  the closed bounded domain  $\mathcal{W}$.
 \end{assumption}

  \begin{assumption}
  \label{ass2}
 The step size $\gamma^{(k)}$ is decreasing and  satisfies  the following
 \begin{align}
     {\sum_{k=1}^{\infty}}\gamma^{(k)}=+\infty\,\,\,\,\,\,\, \text{and} \,\,\,\,\,\,\,   {\sum_{k=1}^{\infty}}\ln k.{(\gamma^{(k)})}^{2}<+\infty.
     \label{StepsizeEq}
 \end{align}
 \end{assumption}
 
 \begin{assumption}[Bounded Gradient]
   \label{ass3}
There exits a constant $D$ such that, $\forall i \in [V]$ and $\forall w\in \mathcal{W}$, we have
\(\left\Vert \nabla F_i(w) \right\Vert^2_2\leq D. \) 
\end{assumption}
 
 The last assumption is actually a result that follows from the functions $F_i$'s being   convex   on a closed bounded subset $\mathcal{W} \subset \mathbb{R}$.  A complementary proof can be found in \cite{ayache2020}.
  
   \section{Random Walk Algorithm}
   \label{sectionRW}

Our objective is to design a random walk on the graph $\mathcal{G}$ algorithm  to learn the optimal model $w^*$.  The algorithm starts uniformly at random at  an initial node in the graph, say $i^{(0)}$,  with an initial model $w^{(0)}$ also sampled uniformly at random from the feasible set $\mathcal{W}$. 
Let $i^{(k)}$ be the node visited (active) at the $k^{th}$ round of the algorithm, $k=0,1,\dots,T$. 
At each round $k$, the active  node $i^{(k)}$ will receive the latest model update $w^{(k)}$  from a neighbor node $i^{(k-1)}$, that was active at the previous round. Then, the model $w^{(k)}$ is updated  
via a gradient descent update using data sampled from the local dataset of node $i^{(k)}$. 


The main question we are after is how to design the transition probabilities defining the random walk, which govern how the RW is sampling the nodes in the graph. In addition to  the explicit objective of learning the model, the random walk will simultaneously learn  information about the heterogeneity  of each   node's  data. 
Therefore, as the random walk progresses, it can  adapt with the information gained on the importance of a given node's data to speed up the convergence. For this reason, we allow the transition probabilities of the random walk to adapt over time (algorithm rounds). 
We denote by $p^{(k)}(i)$ the probability distribution to select the node $i$ to be active at round $k$. We denote by $P^{(k)}$ the transition matrix at time $k$. Therefore, we have $P^{(k)}(i,j)>0$ if $j\sim i$  and $P^{(k)}(i,j)=0$ otherwise. Moreover, we have $p^{(k)}= p^{(0)}P^{(1)}...P^{(k)}$ and $p^{(m,k)}=p^{(0)}P^{(m+1)}...P^{(k)}$.

%


  \section{Node Sampling Strategy}
  \label{sectionNodeSampling}

We aim to design a sampling strategy that mitigates the effect of  heterogeneity on the performance of the  learning algorithm. Such strategy is constrained by an environment with two essential properties to consider: 1) The node sampling is restricted by the topology of the graph where the node to pick next has to be connected to the current active node; 2) No full information about the distributed heterogeneous data is available except what has been learned in the rounds so far and what can be shared among neighbor nodes.



In our algorithm, each node $i$ is sampled (visited) with probability $p^{(k)}(i)$ at round $k$. And the gradient is computed on one data point sampled uniformly among the $n$ local data points at the visited node.
A crucial quantity for our analysis is
the second moment of the unbiased gradient estimate  at round $k$ which is 
is \begin{align}
& \mathbb{E}\left[\left\Vert \hat{\nabla}F_{i^{(k)}}\left(w^{(k)}\right)\right\Vert _{2}^{2}\right] \nonumber
\\ & 
=\underset{i\in[N]}{\sum}\frac{1}{p^{(k)}\left(i\right)}\sum_{j=1}^{n}\frac{1}{n^2}\left\Vert \nabla F_{i}\left(w^{(k)};\xi_{i,j}\right)\right\Vert _{2}^{2}. 
\end{align} 
This quantity affects the convergence rate of the Random Walk SGD algorithm as we show in equation~\eqref{eq:3} (see the Appendix for the details) 
  \begin{align}
 \label{eq:3}
  &  \mathcal{O}\left(\left(\sum_{k=1}^{T}\gamma^{(k)}\right)^{-1}\sum_{k=1}^{T}\mathbb{E}\left[\left\Vert \hat{\nabla}F_{i^{(k)}}\left(w^{(k)}\right)\right\Vert _{2}^{2}\right]\right).
 \end{align}
 
 Thus, the second moment of the gradients' updates 
 imposes a burden on the convergence, especially in heterogeneous data settings where the diversity of the gradients is high. 
 This dependence on the second moment is a common property of  SGD based algorithms and has been well studied in the literature. Variance reduction techniques via importance sampling have been proposed to improve the convergence guarantees \cite{Needell,Zhao2014StochasticOW, Alain2015VarianceRI, bouchard2015online, Papa2015AdaptiveSF}.
 
Our goal is to design the node sampling strategy to approach the optimal probability that minimizes the convergence bound \cite{Zhao2014StochasticOW,Needell,Alain2015VarianceRI} given by  \begin{align*}
& p^{(k)}(i)\propto \sqrt
{g^{(k)}_i \left(w^{(k)}\right)}, \text{ such that }
\\ 
&   g^{(k)}_i\left(w^{(k)}\right)\coloneqq \sum_{\xi_{i,j}\in \mathcal{D}_{i}}\frac{1}{n^2}\left\Vert \nabla F_{i}\left(w^{(k)};\xi_{i,j}\right)\right\Vert _{2}^{2}.
\end{align*}

Note that computing the $g^{(k)}_i$'s is very costly since it requires  computing the gradients related to all the data points owned by the node and its neighbors. Moreover, the $g^{(k)}_i$'s need to be re-computed at the new model $w^{(k)}$ at each iteration.  
Instead, we propose to estimate in each iteration the   $g^{(k)}_i$'s using the already computed gradients for the update step in \eqref{eq:2}. 
Therefore, at each iteration, the random walk has  a double-fold objective: (i) update the model in each iteration and (ii) refine the estimates of the $g_i^{(k)}$'s by adjusting the RW level of exploitation vs.\ exploration using tools from the theory of sleeping multi-armed bandit.


\section{Sleeping Bandit for RW Node Sampling}
 \label{SectionSleepingSampling}

The multi-armed bandit (MAB) problem \cite{Lattimore2020BanditA,bubeck2012regret} is a decision framework that features a set of $N$ arms, where each arm $i \in [N]$ has an unknown   cost $c^{(k)}(i)$ at round $k$. A player selects a sequence of arms $i^{(0)},\,i^{(1)}, ... $ up to the final round $T$ (also called called horizon) of the algorithm. The goal is to design an arm selection strategy to minimize the accumulated regret $R(T)$  over the total number of rounds $T$:
\begin{align}  \text{R}\left(T\right) = \sum_{k=1}^{T}\left(\mathbb{E}\left[c^{(k)}\left(i^{(k)}\right)\right]-\underset{i \in [N]}{\min}\, \mathbb{E}\left[c^{(k)}\left(i\right)\right]\right).\label{regret}
\end{align}
The first term in the regret, $\mathbb{E}\left[c^{(k)}\left(i^{(k)}\right)\right]$, is the average cost of the arm selected by the player. The second term, $\underset{i \in [N]}{\min}\, \mathbb{E}\left[c^{(k)}\left(i\right)\right]$, is the ``best" arm with the minimum cost that could have been selected were the costs   known. The expectation is taken over the selection strategy and the cost randomness.


Our work is based on establishing an analogy between MAB and RW. This allows us to use results from the vast literature on MAB to design the RW in order to speed up the learning process. 
To that end, we think of  each node as an arm, and visiting a node   in the RW as selecting an arm in the MAB problem. 
What is not clear in this analogy is what   the cost of visiting a node and updating the model  is. Based on the discussion in section~\ref{sectionNodeSampling} and the the upper bound in~\eqref{eq:3}, minimizing the accumulated variance of the gradient serves to tighten the convergence guarantees. Thus, we propose the cost of visiting a node $i$, at a given round $k$ in our  Random Walk   algorithm, to be:

\begin{align}  & c^{(k)}(i)=\sum_{\xi_{i,j}\in\mathcal{D}_{i}}\frac{1}{n^{2}}\left\Vert \nabla F_{i}\left(w^{(k)};\xi_{i,j}\right)\right\Vert _{2}^{2}.
\label{cost}
\end{align}




\begin{table}[t]
    \centering
    \begin{tabular}{ | p{0.1\linewidth} | p{0.6\linewidth}|}
       \hline
      MAB           &  RW Learning on Graph     \\
      \hline   \hline
      Arm           &  Node                                \\
         \hline
      Action        &  Select the next node in the RW  \\
         \hline
         
      Cost          & Variance of the local gradient at the selected node as shown in \eqref{cost}\\
         \hline
      Regret        & Gap between the accumulated variance and the minimal variance under the optimal transition probabilities in full information setting as shown in \eqref{regret}\\
       \hline
      
    \end{tabular}
     \vspace{0.2cm}
    \caption{Our proposed Analogy between the sleeping MAB and the Random Walk design problems.  }
    
    \label{tabanalogy}
\end{table}

However, this analogy between ``standard" MAB and RW cannot be fully established here. That is because in MAB any arm can be selected at any time. Whereas in RW   only neighboring nodes can be visited in each iteration. To take into account the graph topology, we consider a variant of the standard MAB called sleeping MAB, where in each iteration only a subset or arms is available (the rest are sleeping)\cite{Kanade2009SleepingEA,Kleinberg2010RegretBF,Saha2020ImprovedSB}.
Moreover, the available nodes to select from are the ones that are   connected to the currently visited node. 
In Table~\ref{tabanalogy}, we summarise this  analogy between MAB and RW.



Within this sleeping multi-armed bandit framework, our goal is to minimize the regret given the available arms and approach the best node sampling strategy denoted by $\pi: 2^{[N]} \rightarrow [N]$, which is a mapping from
a set of available arms $\mathcal{N}^{(k)}$ to a selected arm. The goal is to minimize the  following regret  
\begin{equation}  
\begin{aligned}[b]
& \text{R}\left(T\right)= \\ &
  \sum_{k=1}^{T}\left(\mathbb {E}\left[c^{(k)}\left(i^{(k)}\right)\right]-\min_{\pi}\sum_{k=1}^{T}\mathbb{E}\left[c^{(k)}\left(\pi\left(\mathcal{N}^{(k)}\right)\right)\right]\right),  
  \label{regret}
\end{aligned}
\end{equation}
 where the cost $c^{(k)}$ is defined in \eqref{cost}, the expectation is taken w.r.t. the availabilities and the randomness of the player’s strategy, and $\mathcal{N}^{(k)}$ is the set of available nodes at time $k$ which consists of the neighbors of the currently visited node. Therefore, the regret function is defined as the gap between the local variance of the local gradient estimate implied by the our selection strategy and the minimal variance that requires full information about the local datasets.
 

In \cite{Kanade2009SleepingEA, Kleinberg2010RegretBF,bubeck2012regret}, it was shown that one can achieve a sublinear regret $\mathcal{O}(\sqrt{T})$ for sleeping MAB and it is asymptotically optimal. This  is achieved by applying the EXP3 algorithm. Initially, the algorithm assigns  equal importance to all arms. Then, at every round, the player receives the subset of non-sleeping arms, selects one among them, and observes the outcome of the chosen arm. The player then updates its cost estimation and  keeps track of the empirical probability of the appearance of a given arm in the non-sleeping set. The goal of the player is to balance between exploration and exploitation and gradually shifts to exploitation as the costs estimates become more robust after playing enough rounds.


The multi-armed bandit modeling implies an algorithm design on the random walk that guarantees a sublinear decaying 
of the regret in \eqref{regret}  such that $\lim_{T\rightarrow\infty}\frac{\text{Regret}\left(T\right)}{T}=0$, thus, it approaches  asymptotically the optimal transition scheme of the random walk.

\section{Main Results}
\label{sectionResults}

\begin{algorithm}[t]
\label{Alg1}
   \caption{Sleeping MAB  Random Walk SGD}
\begin{algorithmic}[1]
\STATE{ {\bfseries Input:} {Exploration parameter} $\lambda^{(k)}$. Learning parameter $\eta = \sqrt{\frac{\log N}{NT}}$. Horizon T. Graph $\mathcal{G}(E,V)$.} 

 \STATE{ {\bfseries Initialization:} Initial control weight $q^{(0)}(i) =1 $ $\forall i \in [N]$, Initial model $w^{(0)}$ chosen uniformly at random from $\mathcal{W}$. Starter node $i^{(0)}$ chosen uniformly at random from $[N]$. }

   \FOR{$k=1$ {\bfseries to} $T$}
       \STATE Compute $\ensuremath{P^{(k)}(i^{(k-1)},i) \propto {q^{\left(k-1\right)}\left(i\right)}}  \quad \forall i\in\mathcal{N}\left(i^{\left(k-1\right)}\right)$ and $P^{(k)}(i^{(k-1)},i)=0$ otherwise.
       
    \STATE Choose a neighbor node $i^{(k)} \sim P^{(k)}(i^{(k-1)},\,.)$.
    \STATE Choose $\xi_{i,j}$ uniformly at random from $\mathcal{D}_{i^{(k)}}$ and compute $\hat{\nabla}F_{i^{(k)}}\left(w^{(k)}\right)$.
    
    \STATE Compute the cost estimate $c^{(k)}_{i}$.

    \STATE Update the model using the SGD update in  \eqref{eq:2}.

        \STATE Update the control weight 
        $q^{(k)}(i^{(k)})$ using  ~\eqref{expupdate}.

   \ENDFOR
   
 
\end{algorithmic}
\label{Alg1}
\end{algorithm}


In this   section, we summarize our main technical results.  First, we present the details of our  Sleeping Multi-Armed Bandit Random Walk SGD algorithm in Algorithm~\ref{Alg1}. 
Second, we prove  in Theorem~\ref{maintheorem} that the proposed algorithm has an asymptotically optimal convergence rate. 





\subsection{Algorithm}



Algorithm~\ref{Alg1} leverages the analogy between Sleeping MAB and RW that we established in the previous section to design the  RW learning algorithm. In the literature of Sleeping MAB \cite{Saha2020ImprovedSB}, there are two versions of the EXP3 algorithm based on the availabilities of the arms: dependent versus independent availabilities. The case with dependent availabilities fits our RW model since the graph structure dictates the joint availability of any set of nodes.

In Algorithm~\ref{Alg1}, each node $i$ keeps an accumulated control (importance) value $q^{(k)}(i)$ of the observed cost   up to round $k$. The nodes with higher values will be favored in the selection.

In each round of the algorithm, the active node has to make a decision to select a neighboring node to carry the next update. In order to do that, the active node receives the control value from each neighboring node, and does the selection proportionally to the control values $q^{\left(k-1\right)}\left(i\right)$. 

The selection starts with pure exploration using uniform control values and keeps refining it with time given the observed average cost. The selected node will turn active, samples one of its local data point uniformly at random, performs the update in \eqref{eq:2} and computes the cost estimate based on the local gradient.

Each node $i$ keeps tracks of the empirical estimate $\bar{P}^{(k)}(i)=\frac{1}{k}\sum_{t=1}^{k}\left(P^{(t)}(i^{(t-1)},i)\right)$.
The exploration is implicitly adjusted by a decreasing exploration parameter $\lambda^{(k)}=\sqrt{\frac{2^{N+2}}{k}\ln\left(T\right)}+\frac{2^{N+2}}{3k}\ln\left(T\right)$. At early stage of the training $\lambda^{(k)}$ gives less importance to the observed cost contribution. Lastly the control value is updated as follows
\begin{align}\label{expupdate}q^{\left(k\right)}\left(i^{(k)}\right)=q^{\left(k-1\right)}\left(i^{(k)}\right)\exp\left(-\eta\frac{c_{i}^{(k)}\left(i^{(k)}\right)}{\bar{P}^{\left(k\right)}\left(i^{(k)}\right)+\lambda^{\left(k\right)}}\right).\end{align}

\subsection{Convergence Guarantees} 

\begin{theorem}
\label{maintheorem}
 Under assumptions \ref{ass1}, \ref{ass2} and \ref{ass3}, for a connected graph $\mathcal{G}$,  particularly for $\gamma^{(k)} = \frac{1}{k^q}$ for $\frac{1}{2}<q<1$, the convergence rate of Algorithm \ref{Alg1} is as follows:


\begin{align*}  \mathbb{E}\left[F\left(\bar{w}^{(T)}\right)-F\left(w^{*}\right)\right] \leq\frac{\frac{c.D^{2}}{\ln\left(1/\mu_{\mathcal{G}}\right)}+N.E+\frac{1}{2}R+3C^{*}}{T^{1-q}},
\end{align*}
where,

\begin{align*}
\bar{w}^{(T)}=\frac{\sum_{k=1}^{T}\gamma^{(k)}w^{(k)}}{\sum_{t=1}^{T}\gamma^{(t)}}, & \, C^{*}=\min_{p}\sum_{k=1}^{T}\mathbb{E}_{p}\left[c^{(k)}(i)\right],\\
R=\frac{1}{2}\left\Vert w^{\left(0\right)}-w^{*}\right\Vert _{2}^{2}, & \text{ and }E=\sum_{k=1}^{T}\gamma^{\left(k\right)}\left(\frac{1}{2k}+\frac{1}{\sqrt{k}}\right).
\end{align*}
Moreover, $c$ is a constant function of the convexity constants and the step size and $\mu_{\mathcal{G}}$ is the spectral norm of the transition matrix defined in Section~\ref{sectionproof}. 
\end{theorem}

The first order contribution of Theorem~\ref{maintheorem} is to characterize the convergence rate approaching $\mathcal{O}(\frac{1}{\sqrt{T}})$ of our proposed bandit random walk SGD algorithm. This proves its asymptotic optimality given the lower bound $\mathcal{O}(\frac{1}{\sqrt{T}})$ in \cite{duchi}.  To better understand its significance, one must compare it with other first-order random walk algorithms,   such as uniform node sampling.  For all these algorithms, one can get a similar bound as in Theorem~\ref{maintheorem} with the same constants (depending on the data and graph topology, etc.). The only difference would be in    the constant    $C^{*}$ which is the cumulated variance of the gradients corresponding to the optimal sampling strategy $p^*$. Any other node sampling strategy $p$ will lead to  a higher constant $C$ and a looser bound. Of course, here we are   optimizing the upper bound which we are taking as a  proxy for the actual performance. Our numerical results in Section \ref{sectionVI} substantiate  our theoretical conclusions and show that our proposed algorithm outperform other baselines.

\section{Proof Outline }
\label{sectionproof}

We outline here the different steps needed to establish the result in Theorem \ref{maintheorem}. The details can be found in the Appendix. First,
we state the results on the regret rate of the sleeping multi-armed bandit selection scheme used in Algorithm  \ref{Alg1}. Furthermore, we show that the multi-armed bandit random walk is strongly ergodic which is an essential assumption for the convergence of our algorithm. 


 \begin{lemma}[Regret Rate of Sleeping Multi-Armed Bandit Sampler]
Let $C^* = \min_{p} \sum_{k=1}^{T}\mathbb{E}\left[c^{(k)}(i)\right] $ is the minimal cost if the optimal transition scheme is known.
Under algorithm~\ref{Alg1}, The multi-armed bandit sleeping algorithm approximates the optimal cost  asymptotically as follows 
\begin{align*}
   \lim _{T \rightarrow \infty} \frac{1}{T} \left( \sum_{k=1}^{T}\mathbb{E}\left[c^{(k)}(i^{(k)})\right]-3C^*\right) \leq 0,
\end{align*}

 \end{lemma} The proof follows the multiplicative weight approach for EXP3 algorithms introduced in \cite{auer2002nonstochastic}.



Next, we state the definition on Strongly Ergodic Non-Homogeneous Random Walk \cite{huang1977non}. The sleeping multi-armed bandit algorithm guarantees that this property applies on the sequence of employed transition, which in the end guarantees the convergence stated in Theorem \ref{maintheorem}. 

\begin{defn}[Strongly Ergodic Non-Homogeneous Random Walk \cite{huang1977non}]
\label{definition}
A non-homogeneous random walk, with uniform starting distribution $p^{(0)}$, is called strongly ergodic if there exists a vector $p^*$ such that for all $m \geq 0$, $$\lim _{k \rightarrow \infty}  \left \Vert p^{(m,k)} - p^*  \right \Vert =0.$$
\end{defn}

Lastly, we present the result on the rate of convergence of the transition probability distribution. 

\begin{prop}[Convergence Non-Homogeneous Strongly Ergodic Random Walk]

The non-homogeneous random walk in Algorithm~\ref{Alg1} is  strongly ergodic.  
Thus, it exists  a stochastic matrix $P$ such that $\lim_{k\rightarrow\infty}\left\Vert P^{(k)}-P\right\Vert =0$. Moreover, it exists  a stochastic matrix $Q$ such that $\left\Vert P^{k}-Q\right\Vert \leq c\beta_2^{k}$, where $\beta_{1}=1>\beta_{2}\geq...\geq\beta_{N}$ are  the eigenvalues of  the matrix $P$. Moreover, it exists a function $g(k) = \mathcal{O}(\sqrt{k})$ such that

$\lim_{k\rightarrow\infty}\min{\left\{ 1/\mu_{\mathcal{G}}^{k},\:g\left(k\right)\right\} }\left\Vert P^{(0,\,k)}-Q\right\Vert =0,$
where  $1<1/\mu_{\mathcal{G}} < \sqrt{1/\beta_2} .$

\label{prop}
\end{prop}






 \section{Simulations}

 \label{sectionVI}
 
 In this section, we present the numerical performance of our proposed   Multi-Armed Bandit Random Walk (RW) SGD algorithm described in Algorithm \ref{Alg1}. 
 
 
 \subsection{Baseline Algorithms}

We compare the performance of our algorithm to three baselines,   namely: (1) Uniform Random Walk,  (2)   Static Weighted Random Walk, and (3) Adaptive Weighted Random Walk.

 \paragraph{The Uniform Random Walk} This algorithm
  assigns equal importance to all nodes in the network  \cite{Johansson2007ASP} imitating uniform sampling in centralized SGD.  We implement the  Metropolis Hasting (MH) decision rule to design the transition probabilities, so the random walk converges to a uniform stationary. The MH rule can be described as follows:
 \begin{enumerate}
 \item At the  $k^{th}$ step of the random walk, the active node $i^{(k)}$ selects uniformly at random one of its neighbors, say $j$, as a candidate to be the next active node.  This selection gets accepted with probability
 \begin{align*}a_u\left(i^{(k)},\,j\right)=\min\left(1,\,\frac{\deg\left(i^{(k)}\right)}{\deg\left(j\right)}\right).\end{align*}
 Upon the acceptance, we have $i^{(k+1)}=j$.
 \item Otherwise, if the candidate node gets rejected,  the random walk stays at the same node, i.e.,  $i^{(k+1)} = i^{(k)}$. 
  \end{enumerate}

 \paragraph{Static Weighted Random Walk} This algorithm assigns a static importance metric to each node that is proportional to the gradient-Lipschitz constant of the local loss function \cite{Ayache2019RandomWG, Needell}. 
 The random walk is designed by the MH again with a stationary distribution that is proportional to the local gradient-Lipschitz constants. In order to achieve that stationary, the probability of   acceptance looks as follows:

\begin{align}a_{w}\left(i^{(k)},\,j\right)=\min\left(1,\,\frac{L_{j}}{L_{i^{(k)}}}\frac{\deg\left(i^{(k)}\right)}{\deg\left(j\right)}\right).\label{Metropolis}\end{align}

 \paragraph{Adaptive Weighted
Random Walk}  In this algorithm, we adapted the importance sampling scheme that is used in  \cite{Zhao2014StochasticOW,Stich2017SafeAI} for centralized settings.  The importance is computed for each node $i$ as the average of of gradients computed so far at that node which is at time $k$, $\frac{\sum_{t=1:i^{(t)}=i}^{k}\left\Vert \hat{\nabla}F_{i}\left(w^{(t)}\right)\right\Vert _{2}^{2}}{n_{i,k}}$, for $n_{i,k} $ is the total number of rounds when node $i$ has been active up to $k$. We call it the pure Exploitation scheme. 
 
 \subsection{Datasets and Comparison}
 Our simulations are run on both synthetic dataset and on real benchmarks to confirm our theoretical results. They show that a bandit based random walk in decentralized learning consistently outperforms existing random walk baselines that uses static or exploitation based importance estimation.
 


\vspace{0.2cm}

\noindent {\bf{Synthetic Data.}} For each node $i$, we sample the dataset $\mathcal{D}_i$ from a  a normal distribution with $\mathcal{N}\left([\mu_{i,1},\mu_{i,2}], 
\sigma \mathbb{I}_{2}\right)$. We assigned manually a label to each node dataset such that half the nodes has the label $y_i = 1$ and the other half has the $y_i = - 1$. 
We run our simulations on  an Expander graph  which known to be sparse {(The Margulis-Gabber-Galil graph). \footnote{We call the generator function of the Python library NetworkX v2.8 \cite{SciPyProceedings_11}. }}.

\vspace{0.2cm}

 \noindent {\bf{MNIST dataset and Fashion-MNIST.}} We run experiments on the MNIST dataset and the Fashion-MNIST to train a multi-class logistic regression model. We divide the data among the nodes as follows: for a level $s\%$ of similarity,  each client has $s\%$  of its local dataset drawn i.i.d. from a shared pool of data. Another non-public pool of the data is sorted with respect to label and  partitioned into non-overlapping chunks. Each node is assigned a different chunk that consists of its remaining $\left(100-s\right) \%$ data \cite{karimireddy2020scaffold}.

Table~\ref{tab2} and Table~\ref{tab4} report the results for different level of similarity from fully homogeneous to fully heterogeneous to  highlight how different similarity levels affect the convergence of our algorithm vs. the oblivious uniform algorithm. Table~\ref{tab2} is on the MNIST dataset. The gap between both algorithms becomes wider once the system turns more heterogeneous. Table~\ref{tab4} is on the Fashion-MNIST dataset. The gap between both algorithms becomes more significant as the system turns into more heterogeneous state \footnote{The fact that the ratio between the number of iterations of the two algorithms is constant (roughly 2) is an  artifact of the data and is not reproducible for other datasets. See for example Table~\ref{tab4} which is on the Fashion-MNIST dataset that shows  an increasing ratio  with decreasing similarity. }.

\begin{figure}[!htb]
    \centering
    \includegraphics[scale=0.4    ]{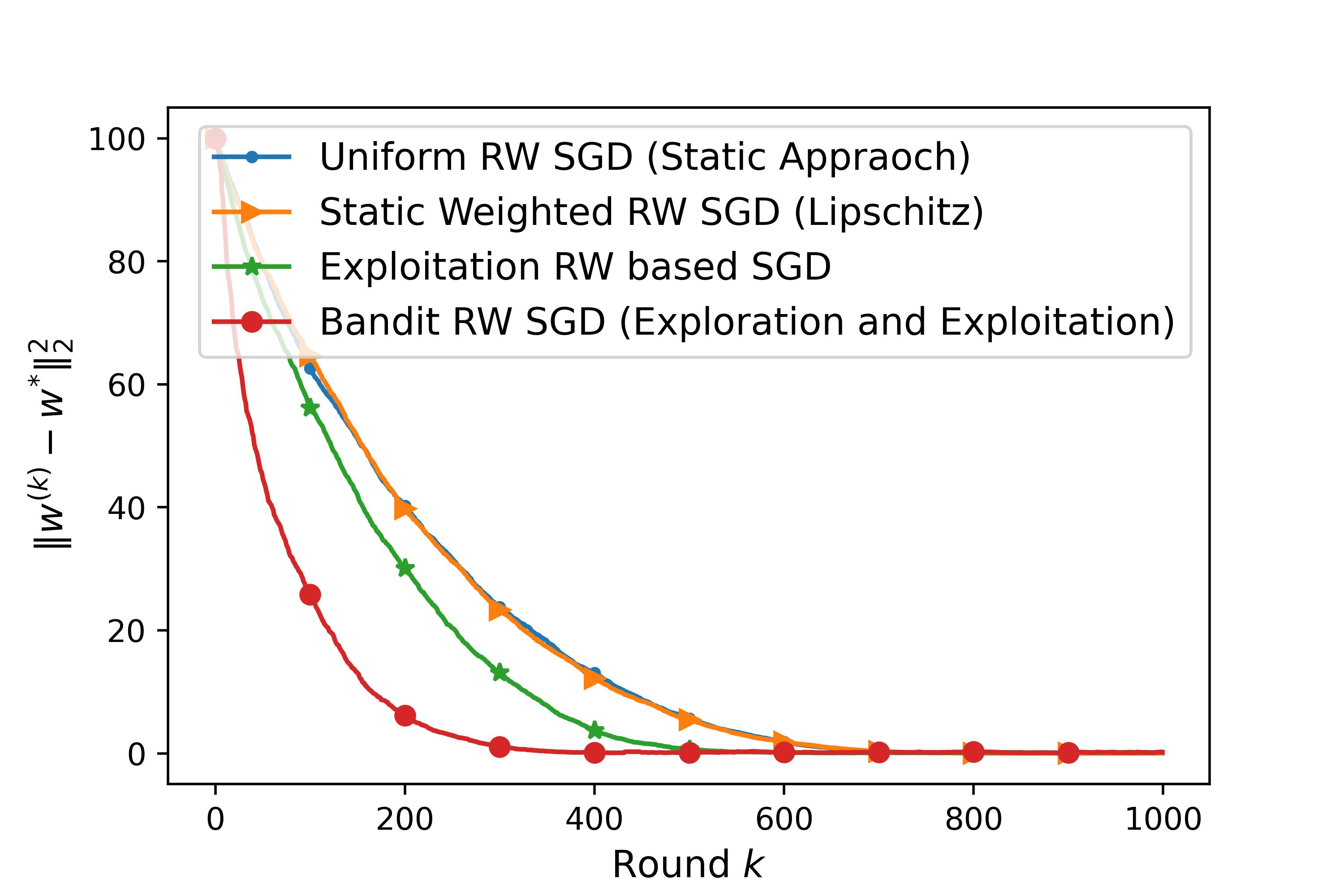}
    \caption{ Classification model trained on a Synthetic dataset  distributed over an Expander graph of $100$ nodes.}
    \label{fig:my_label}
\end{figure}
\begin{figure}[!htb]
    \centering
    \includegraphics[scale=0.4]{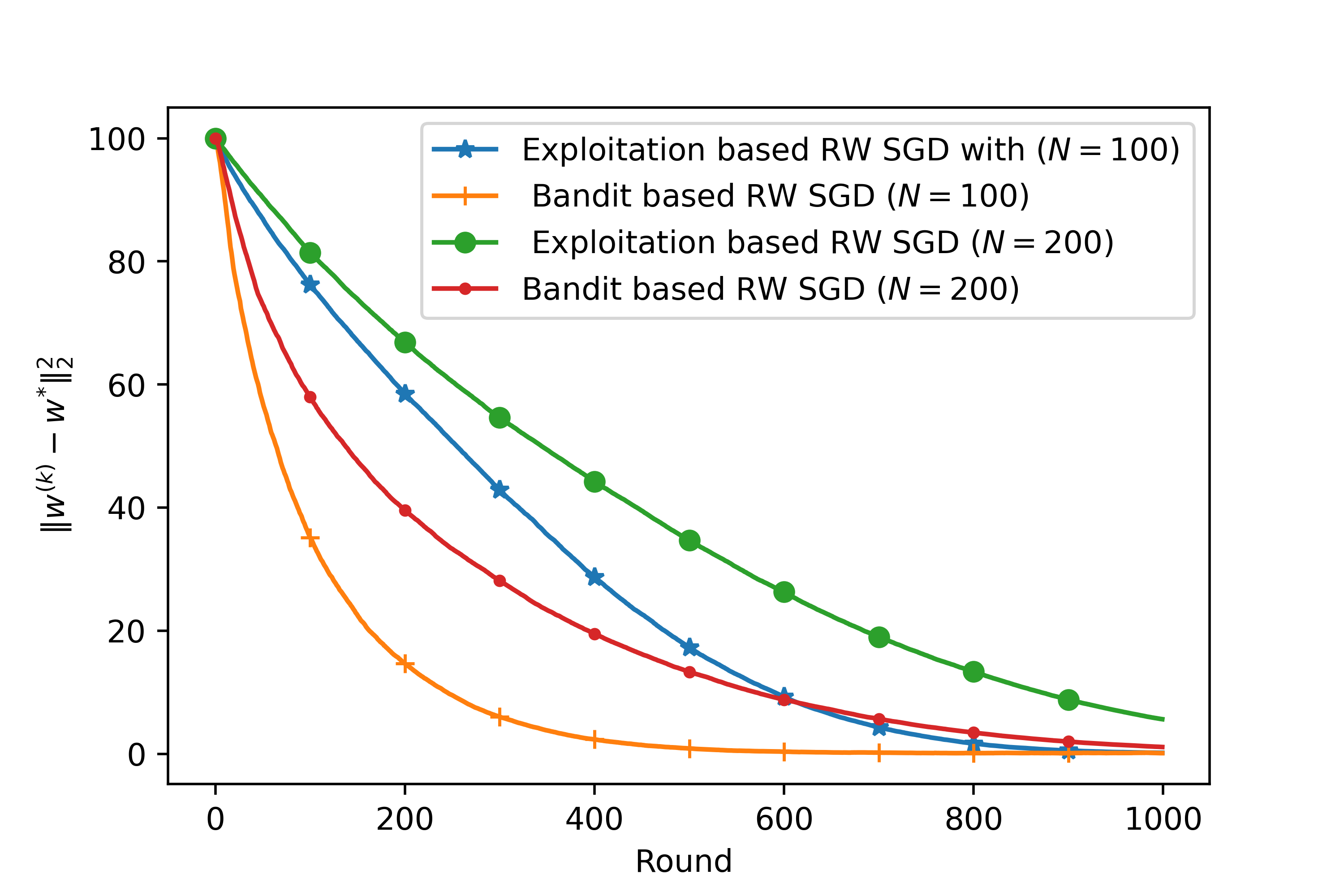}
    \caption{Classification model trained on the Synthetic dataset. The same dataset is been distributed over a Expander graph of size  $100$ nodes and $200$ nodes.}
    \label{fig:my_label}
\end{figure}
\begin{figure}[!htb]
    \centering
    \includegraphics[scale=0.4    ]{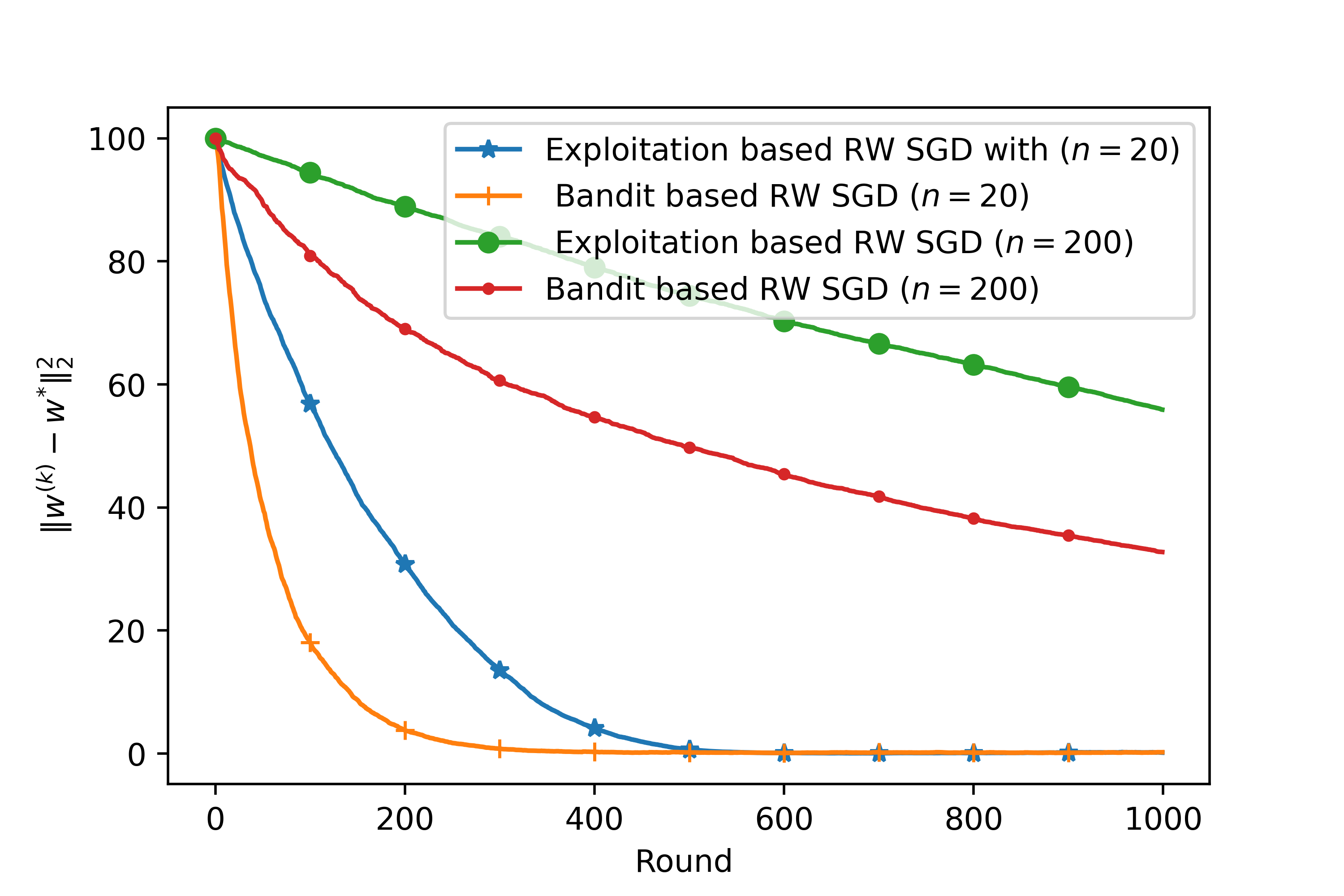}
    \caption{Classification model trained on the Synthetic dataset. The local dataset size has been augmented from  $20$ to $200$ in a Expander graph of size $100$.}
    \label{fig:my_label}
\end{figure}
\begin{figure}[!htb]
    \centering
    \includegraphics[scale=0.4    ]{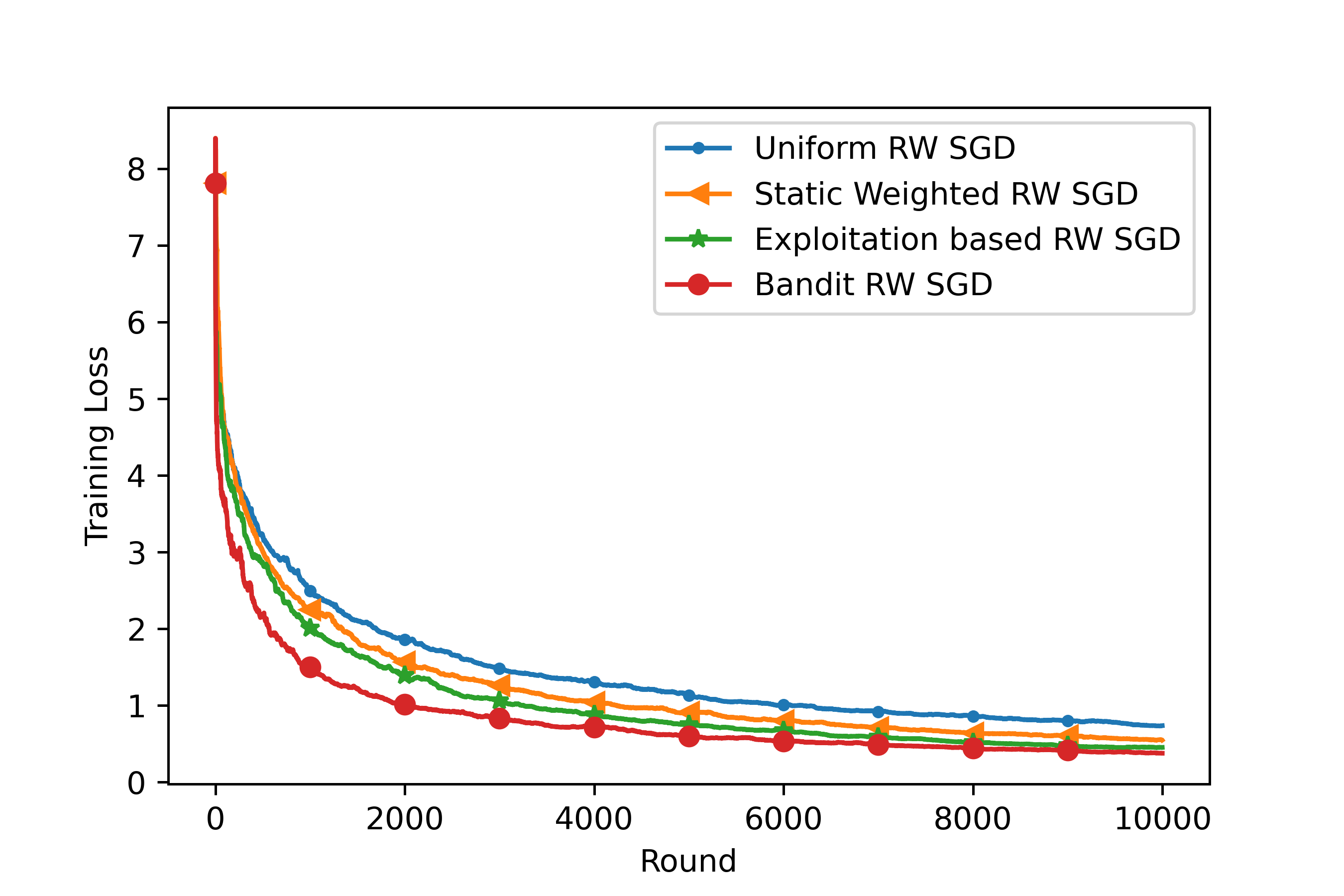}
    \caption{Multi-class MNIST dataset of 10 classes distributed over $100$ nodes with $0\%$ similarity on Expander graph. }
    \label{fig:my_label}
\end{figure}
 \begin{table}[!htb]
   \centering
    \begin{tabular}{c|c|c|c}
   Similarity & $0\%$ & $10\%$ & $100\%$ \\
      Uniform RW SGD & $202$ & $87$ & $61$ \\
       Bandit RW SGD  & $138$ & $44$ & $34$  
    \end{tabular}
     \vspace{0.2cm}
    \caption{Number of rounds to reach 0.45 test accuracy for
logistic regression on MNIST as we vary the level of similarity. Bandit RW SGD is consistently faster than Uniform RW SGD.}
    \label{tab2}
\end{table}

       \begin{table}[!htb]
   \centering
    \begin{tabular}{c|c|c|c}
   Similarity & $0\%$& 10\% &  $100\%$ \\
      Uniform RW SGD & $124$ & 90 & $66$ \\
       Bandit RW SGD  & $82$ & 69 &$54$  
    \end{tabular}
     \vspace{0.2cm}
    \caption{Number of rounds to reach 0.45 test accuracy for
logistic regression on FMNIST as we vary the level of similarity. Bandit RW SGD is consistently faster than Uniform RW SGD.}
    \label{tab4}
\end{table}

\begin{table}[!htb]
    \centering
    \begin{tabular}{c|c|c|c}
   Probability of Connectivity & $0.1$ & $0.5$ & $0.8$  \\ 
      Uniform RW SGD & $202$ & $128$ & $94$ \\
       Bandit RW SGD  & $138$ & $103$ & $90$  
    \end{tabular}
     \vspace{0.2cm}
    \caption{Number of rounds to reach 0.45 test accuracy for
logistic regression on MNIST as we vary the graph connectivity parameter. Bandit RW SGD is consistently faster than Uniform RW SGD. As we decrease the probability of
connectivity, the increase in the number of rounds is less significant for the bandit algorithm.}
    \label{tab3}
\end{table}

To elaborate on how our algorithm performs given the graph structure, we consider multiple scenarios of simulations where we use an Erdos-Renyi with different probabilities of connectivity that go from sparser to denser.
Table~\ref{tab3} measures the performance given the probability of connectivity in the graph.

\appendix

\textbf{Optimal Sampling.} Here is a sketched proof on the optimal sampling. Consider the   optimization problem in hand which can be formulated as follows:
\begin{align*} & \min\underset{i\in[N]}{\sum}\frac{g_{i}^{(k)}\left(w^{(k)}\right)}{p^{(k)}\left(i\right)},\\
 & \text{such that\ensuremath{\quad}}{p^{(k)}\left(i\right)}\in(0,\,1)\,\forall\,i,\text{ and }\sum_{i}{p^{(k)}\left(i\right)}=1.
\end{align*}
The optimality conditions of the Lagrangian expression of the problem above gives the following:
    \begin{align*}-\frac{g_{i}^{(k)}\left(w^{(k)}\right)}{(p^{(k)}\left(i\right))^2}+\eta=0\text{ and }\sum_{i}{p^{(k)}\left(i\right)}=1,\end{align*}
where $\eta$ is the Lagrange multiplier.
Thus, by a simple algebraic manipulation, we get $p^{(k)}\left(i\right)
=\frac{\sqrt{g_{i}^{(k)}\left(w^{(k)}\right)}}{\sum_{j}\sqrt{g_{j}^{(k)}\left(w^{(k)}\right)}}.$
 \begin{defn}
The local loss function $F_i$ for each node $i \in V$  has an $L_i$-Lipschitz continuous gradient; that is, any $w$,\,$w' \in \mathcal{W}$, there exists a constant $L_i>0 $ such that
\begin{align*}
  \left\Vert \nabla f_i\left(w\right)-\nabla f_i\left(w'\right)\right\Vert _{2}\leq L_{i}\left\Vert w-w'\right\Vert_2.  
\end{align*}
 \end{defn}
\begin{lemma}
\label{lemmaBound}
Under assumptions~\ref{ass1},  ~\ref{ass2} and ~\ref{ass3}, the Random Walk SGD algorithm, that uses the update in equation~3 for transition matrix $P$, has the following rate of convergence.

\begin{align*}
&\mathbb{E}\left[F\left(w^{(T)}\right)-F\left(w^{*}\right)\right] 
\\ & \qquad \leq \mathcal{O}\left(\left(\sum_{k=1}^{T}\gamma^{(k)}\right)^{-1}\sum_{k=1}^{T}\mathbb{E}\left[\left\Vert \hat{\nabla}F_{i^{(k)}}\left(w^{(k)}\right)\right\Vert _{2}^{2}\right]\right).\end{align*}

\end{lemma}


In order to prove Lemma~\ref{lemmaBound} and Theorem~\ref{maintheorem}, we present some technical results that we use in the proof. The proof techniques are essentially inspired by the work of \cite{ayache2020} and theya are adapted to the assumptions and setting of this work.

\begin{lemma} [Convexity and Lipschitzness]
\label{Lipschitz}
If $F_i$ is a convex function on an open subset $\Omega \subseteq \mathbb{R}$, then for a closed bounded subset $\mathcal{W} \subset \Omega$, there exists a constant $D_i\geq0$, such that, for any $w_1,\,w_2 \in \mathcal{W}$,
\begin{align*}
\left|F_{i}\left(w_{1}\right)-F_{i}\left(w_{2}\right)\right|\leq D_i\left\Vert w_1-w_2\right\Vert_2.   \end{align*}

We define $D=\underset{i\in V}{\sup } D_i$. Therefore,
$$\left|F_{i}\left(w_{1}\right)-F_{i}\left(w_{2}\right)\right|\leq D\left\Vert w_1-w_2\right\Vert_2. $$
\end{lemma}
\noindent A proof for Lemma~\ref{Lipschitz} can be found in \cite{convex}.

\begin{cor}[Boundedness of the gradient]
If $F_i$ is a convex function on  $ \mathbb{R}$, then for a closed bounded subset $\mathcal{W} \subset \mathbb{R}$,
\(\left\Vert \nabla F_i(w) \right\Vert_2\leq D, \,\,\,\,\, \forall w\in \mathcal{W}.\)
\begin{proof}
Taking $v = w+\nabla F_i (w)$, 
\begin{align*}D\left\Vert \nabla F_{i}(w)\right\Vert _{2} &=D\left\Vert v-w\right\Vert _{2}
\\ &  \overset{(a)}{\geq}|F_{i}(v)-F_{i}(w)| 
\\ &  \overset{(b)}{\geq}\left\langle \nabla F_{i}\left(w\right),\nabla F_{i}\left(w\right)\right\rangle 
\\ & =\left\Vert \nabla F_{i}(w)\right\Vert _{2}^{2}.
\end{align*}
(a) follows Lemma~\ref{Lipschitz} and (b) follows from $F_i$ being convex.
\end{proof}
\label{cor1}
\end{cor}
Now, we present the steps of the the proof:
\begin{align*}&\left\Vert w^{(k+1)}-w^{*}\right\Vert _{2}^{2} \\ & \quad =\left\Vert {\bf {\bf {\Pi}}_{\mathcal{W}}}\left(w^{(k)}-\gamma^{(k)}\hat{\nabla}{F}_{i^{(k)}}\left(w^{(k)}\right)\right)-{\bf {\bf {\Pi}}_{\mathcal{W}}}F\left(w^{*}\right)\right\Vert _{2}^{2}\\
 & \quad \overset{\left(a\right)}{\leq}\left\Vert w^{(k)}-\gamma^{(k)}\hat{\nabla}{F}_{i^{(k)}}\left(w^{(k)}\right)-w^{*}\right\Vert _{2}^{2}\\
 & \quad =\left\Vert w^{(k)}-w^{*}\right\Vert _{2}^{2}-2\gamma^{(k)}\left\langle w^{(k)}-w^{*},\hat{\nabla}{F}_{i^{(k)}}\left(w^{(k)}\right)\right\rangle \\
 & \quad \,\,\,\,\,+\left(\gamma^{(k)}\right){}^{2}\left\Vert \hat{\nabla}{F}_{i^{(k)}}\left(w^{(k)}\right)\right\Vert _{2}^{2}
\end{align*}

\noindent $\left(a\right)$ follows from $\mathcal{W}$ being a convex closed set, so one can apply nonexpansivity theorem in \cite {convex}.

\noindent For the next we use the convexity of $F_i$,
\begin{equation}
    \begin{aligned}
    \label{eqbound}\left\Vert w^{(k+1)}-w^{*}\right\Vert _{2}^{2}  &  \leq\left\Vert w^{(k)}-w^{*}\right\Vert _{2}^{2} \\ &  \quad -2\gamma^{(k)}\left(F_{i^{(k)}}\left(w^{(k)}\right)-F_{i^{(k)}}\left(w^{*}\right)\right)  \\ & \quad +{(\gamma^{(k)})}\left\Vert \hat{\nabla}F_{i^{(k)}}\left(w^{(k)}\right)\right\Vert _{2}^{2}.
    \end{aligned}
\end{equation}

\noindent Re-arranging the above equation gives

\begin{align}\gamma^{(k)}&\left(F_{i^{(k)}}\left(w^{(k)}\right)-F_{i^{(k)}}\left(w^{*}\right)\right)  \nonumber  \\ &  \leq\frac{1}{2}\left(\left\Vert w^{(k)}-w^{*}\right\Vert _{2}^{2}-\left\Vert w^{(k+1)}-w^{*}\right\Vert _{2}^{2}\right)   \nonumber \\
 & \quad+\frac{{(\gamma^{(k)})}^{2}}{2}\left\Vert \hat{\nabla}F_{i^{(k)}}\left(w^{(k)}\right)\right\Vert _{2}^{2}. 
 \label{eq:8} 
\end{align}

\noindent Summing \eqref{eq:8} over $k$ and using Assumption and the boundness of $\mathcal{W}$,
\begin{align}&{\sum}_{k}\gamma^{(k)}\left(F_{i^{(k)}}\left(w^{(k)}\right)-F_{i^{(k)}}\left(w^{*}\right)\right)  \nonumber \\ & \quad \leq\frac{1}{2}\left\Vert w^{(0)}-w^{*}\right\Vert _{2}^{2}+{\sum}_{k}{(\gamma^{(k)})}^{2}\left\Vert \nabla F_{i^{(k)}}\left(w^{(k)}\right)\right\Vert _{2}^{2}.\label{eqn}
\end{align}

Next we give some results we need on the  Markov chain. We denote by $\mu$  the stationary distribution, $P$ is the transition matrix and $P^k$ is the $k^{th}$ power of matrix P. We refer to   $i^{th}$ row of a matrix $P$ by $P(i,:)$.

\begin{lemma} [Convergence of Markov Chain \cite{Levin}] 
Assume the graph $\mathcal{G}$ is connected with self-loop, therefore a random walk is aperiodic and irreducible,   we have
\begin{align*}
    \underset{i}{\max}\left\Vert \mu-P^{k}\left(i,:\right)\right\Vert \leq C\lambda_{P}^{(k)} 
\end{align*}
 for $k>K_p$, where
$K_P$ is a constant that depends and $\lambda_{P}$ and $\lambda_2(P)$ and $C$ is a constant that depends on the Jordan canonical form of $P$.

\end{lemma} 
\begin{cor}Using the previous lemma, we get 
\begin{align*}
   \underset{i}{\max}\left\Vert \mu-P^{T_{k}}\left(i,:\right)\right\Vert \leq C\lambda_{P}^{T_{k}}\leq\frac{1}{2k} 
\end{align*}
\text{for} {$T^{(k)} = \min \{k, \max\{\frac{\ln({2Ck})}{\ln({1/\lambda_{P}})},K_P\} \}.$}
\label{LargeT}
\end{cor}

Here, we state the next corollary on the convergence of the random walk.

\begin{align*} & \gamma^{(k)}\mathbb{E}\left[F_{j^{(k)}}\left(w^{\left(k-T^{\left(k\right)}\right)}\right)-F_{j^{(k)}}\left(w^{(k)}\right)\right]  \\  & \quad \overset{(a)}{\leq}D\gamma^{(k)}\mathbb{E}\left\Vert w^{\left(k-T^{\left(k\right)}\right)}-w^{(k)}\right\Vert \\
 &  \quad \overset{(b)}{\leq}D\gamma^{(k)}\mathbb{E}\left(\sum_{n=k-T^{\left(k\right)}}^{k-1}\left\Vert w^{\left(n+1\right)}-w^{\left(n\right)}\right\Vert \right)\\
 & \quad \overset{(c)}{\leq}D\gamma^{(k)}\sum_{n=k-T^{\left(k\right)}}^{k-1}\mathbb{E}\left(\left\Vert w^{\left(n+1\right)}-w^{\left(n\right)}\right\Vert \right)\\
 & \quad  \overset{(d)}{\leq}D^{2}\gamma^{(k)}\sum_{n=k-T^{\left(k\right)}}^{k-1}\gamma^{\left(n\right)}\\
 & \quad \overset{(e)}{\leq}\frac{D^{2}}{2}\sum_{n=k-T^{\left(k\right)}}^{k-1}\left({(\gamma^{\left(n\right)})}^{2}+{(\gamma^{\left(k\right)})}^{2}\right)\\
 & \quad \overset{}{\leq}\frac{D^{2}}{2}T^{\left(k\right)}{(\gamma^{\left(k\right)})}^{2}+\frac{D^{2}}{2}\sum_{n=k-T_{k}}^{k-1}{(\gamma^{\left(n\right)})}^{2}.
\end{align*}

(a) follows from Lemma~\ref{Lipschitz}, (b) using triangle inequality, (c) using linearity of expectation and (d) follows from the Cauchy–Schwarz inequality.

\noindent Now taking the summation over $k$:
 \begin{align} &  \sum_{k}\gamma^{(k)}\mathbb{E}  \left[F_{j^{(k)}}\left(w^{(k-T^{(k)})}\right)-F_{j^{(k)}}\left(w^{(k)}\right)\right]
\nonumber \\ &  \leq\sum_{k}\frac{D^{2}}{2}T_k{(\gamma^{(k)})}^{2}+\frac{D^{2}}{2}\sum_{k}\sum_{n=k-T_k}^{k-1}{(\gamma^{(n)})}^{2}.
 \label{bound1}
 \end{align}
By simply using the assumption on the step size summability, the result is as follows:
      \begin{align}
      \label{FirstEq}
    \sum_{k=K}^{\infty}\sum_{n=k-T^{(k)}}^{k-1}\left(\gamma^{\left(n\right)}\right)^{2}
   & \leq\sum_{k=K}^{\infty}T^{\left(k\right)}\left(\gamma^{(k)}\right)^{2}
    \nonumber \\ &  \leq\frac{2}{\ln\left(1/\lambda_{P}\right)}\sum_{k=K}^{\infty}\ln k.\left(\gamma^{(k)}\right)^{2} \nonumber \\ & <\infty.  \end{align}
Now, we compute the following lower bound:
\begin{align} & \mathbb{E}_{j^{(k)}}\left[F_{j^{(k)}}\left(w^{(k-T^{\left(k\right)})}\right)-F_{j^{(k)}}\left(w^{*}\right)|\,X_{0},\,X_{1},\,...,\,X_{k-T^{\left(k\right)}}\right]\nonumber \\
 & \overset{}{=}\sum_{i=1}^{N}\left(F_{i}\left(w^{(k-T^{\left(k\right)})}\right)-F_{i}\left(w^{*}\right)\right)P\left(j^{(k)}=i\,|\,X_{0},\,X_{1},\,...,\,X_{k-T^{\left(k\right)}}\right)\nonumber \\
 & \overset{(a)}{=}\sum_{i=1}^{N}\left(F_{i}\left(w^{(k-T^{\left(k\right)})}\right)-F_{i}\left(w^{*}\right)\right)P\left(j^{(k)}=i\,|\,X_{k-T^{\left(k\right)}}\right)\nonumber \\
 & \overset{}{=}\sum_{i=1}^{N}\left(F_{i}\left(w^{(k-T^{\left(k\right)})}\right)-F_{i}\left(w^{*}\right)\right)P^{T^{(k)}}\left[X_{k-T^{\left(k\right)}}\,|\,j^{(k)}=i\right]\nonumber \\
 & \overset{(b)}{\geq}\left(F\left(w^{(k-T^{\left(k\right)})}\right)-F\left(w^{*}\right)\right)-\frac{N}{2k}.
 \label{eq18}
\end{align}

(a) using Markov property and (b) using Lemma 1 in \cite{MCGD}. \\

Therefore,
\begin{align*}
&   F\left(w^{(k-T^{\left(k\right)})}\right)-F\left(w^{*}\right)  \\ & \leq\frac{N}{2k}+ \mathbb{E}_{j^{(k)}}\left[F_{j^{(k)}}\left(w^{(k-T^{\left(k\right)})}\right)-F_{j^{(k)}}\left(w^{*}\right)|\,X_{0},\,...,\,X_{k-T^{\left(k\right)}}\right]
 \end{align*}
  
\begin{align*} &   \gamma^{\left(k\right)}\mathbb{E}\left[F\left(w^{(k-T^{\left(k\right)})}\right)-F\left(w^{*}\right)\right]
 \\ & \leq\frac{N\gamma^{\left(k\right)}}{2k}+  \gamma^{(k)}\mathbb{E}\left[F_{j^{(k)}}\left(w^{\left(k-T^{\left(k\right)}\right)}\right)-F_{j^{(k)}}\left(w^{*)}\right)\right]
 \end{align*}
  
\begin{align*}
& \sum_{k}\gamma^{\left(k\right)}\mathbb{E}\left[F\left(w^{(k-T^{\left(k\right)})}\right)-F\left(w^{*}\right)\right] \\	   & \leq \sum_{k}\frac{N\gamma^{\left(k\right)}}{2k} \\ & +\sum_{k}\gamma^{(k)}\mathbb{E}\left[F_{j^{(k)}}\left(w^{\left(k-T^{\left(k\right)}\right)}\right)-F_{j^{(k)}}\left(w^{*)}\right)\right]
 \\&	\leq\sum_{k}\frac{N\gamma^{\left(k\right)}}{2k}+\frac{1}{2}\left\Vert w^{(0)}-w^{*}\right\Vert _{2}^{2} \\ & +{\sum}_{k}{(\gamma^{(k)})}^{2} \, \mathbb{E} \left[\left\Vert \nabla F_{i^{(k)}}\left(w^{*}\right)\right\Vert _{2}^{2} \right]
\end{align*}

Next, we get a bound on 

$$  \sum_{k}\gamma^{\left(k\right)}\mathbb{E}\left[F\left(w^{(k)}\right)-F\left(w^{(k-T^{\left(k\right)})}\right)\right].$$
\begin{align}& \gamma^{(k)}\mathbb{E}\left[F\left(w^{(k)}\right)-F\left(w^{\left(k-T_{k}\right)}\right)\right] \nonumber \\ & \overset{(a)}{\leq}ND\gamma^{(k)}\mathbb{E}\left\Vert w^{(k)}-w^{\left(k-T_{k}\right)}\right\Vert \nonumber \\
 & \overset{(b)}{\leq}ND\gamma^{(k)}\mathbb{E}\left(\sum_{n=k-T_{k}}^{k-1}\left\Vert w^{\left(n+1\right)}-w^{\left(n\right)}\right\Vert \right)\nonumber \\
 & \overset{(c)}{\leq}ND\gamma^{(k)}\sum_{n=k-T_{k}}^{k-1}\mathbb{E}\left(\left\Vert w^{\left(n+1\right)}-w^{\left(n\right)}\right\Vert \right)\nonumber \\
 & \overset{(d)}{\leq}ND^{2}\gamma^{(k)}\sum_{n=k-T_{k}}^{k-1}\gamma^{\left(n\right)}\nonumber \\
 & \overset{(e)}{\leq}\frac{ND^{2}}{2}\sum_{n=k-T_{k}}^{k-1}\left({(\gamma^{\left(n\right)})}^{2}+{(\gamma^{\left(k\right)})}^{2}\right)\nonumber \\
 & \overset{}{\leq}\frac{ND^{2}}{2}T_{k}{(\gamma^{\left(k\right)})}^{2}+\frac{ND^{2}}{2}\sum_{n=k-T_{k}}^{k-1}{(\gamma^{\left(n\right)})}^{2}.
\end{align}

(a) follows from Lemma 4, (b) using triangle inequality, (c) using linearity of expectation, (d) follows Corollary 1 and (e) follows from the Cauchy–Schwarz inequality.
The upper bound summability over $k$ follows from previous discussion in equation \eqref{FirstEq}.

\noindent Combining with the results in \eqref{eqn} and \eqref{FirstEq} , we get 

\begin{align*}
& \mathbb{E}\left[F\left(w^{(k)}\right)-F\left(w^{*}\right)\right] \\ & \leq\frac{\sum_{k=1}^{T}\frac{N\gamma^{\left(k\right)}}{2k}+\frac{C.D^{2}}{\ln\left(1/\lambda_{P}\right)}
 +\frac{1}{2}\left\Vert w^{(0)}-w^{*}\right\Vert _{2}^{2}}{\sum_{k=1}^{T}\gamma^{(k)}}
 \\ & \quad +\frac{{(\gamma^{(0)})}^{2}\sum_{k=1}^{T}\mathbb{E}\left[\left\Vert \hat{\nabla}F_{i^{(k)}}\left(w^{(k)}\right)\right\Vert _{2}^{2}\right]}{\sum_{k=1}^{T}\gamma^{(k)}}.\end{align*}


 By this step we proved Lemma~\ref{lemmaBound}. Next we present the essential technical results to use in the proof of Theorem~$1$.
\begin{prop}
\label{convergence}[Sleeping multi-armed bandit convergence \cite{Kleinberg2010RegretBF}]
The sleeping multi-armed bandit sampling scheme under adversarial availabilities guarantees the following:
$\left\Vert P^{(k)}-P\right\Vert \leq\mathcal{O}(\frac{1}{\sqrt{k}})$. Therefore, using Definition~1, the random walk with transition matrices $P^{(k)}$ is strongly ergodic. 

\end{prop}

We state next Lemma of \cite{huang1977non} about the convergence of strongly ergodic random walk.

\begin{lemma}[Theorem II.7 in \cite{huang1977non}]
Given strongly ergodic non-homogenous transition matrices $P^{(k)}$ with a stochastic matrix $P$ such that $\lim_{k\rightarrow\infty}g(2k)\left\Vert P^{(k)}-P\right\Vert =0$, given $Q$ such $\left\Vert P^{k}-Q\right\Vert \leq c\beta_2^{k}$,  then $\lim_{k\rightarrow\infty}\min{\left\{ (1/\mu)^{k},\:g\left(k\right)\right\} }\left\Vert P^{(0,\,k)}-Q\right\Vert =0$, where 
 $1<1/\mu < \sqrt{1/\beta_2} .$

\end{lemma}

Using the previous lemma, we get 
\begin{align}\underset{i}{\max}\left\Vert \mu -P^{(0,\,T_{k})}\left(i,:\right)\right\Vert 
\leq\mathcal{O}
\left(\frac{1}{2k}+\frac{1}{\sqrt{k}}\right)\end{align}

for {$T_k = \min \left\{k, \max\left\{\frac{\ln({2Ck})}{\ln({1/\lambda})} ,K_P\right\} \right\}.$}
Therefore,

\begin{align} & \mathbb{E}_{j^{(k)}}\left[F_{j^{(k)}}\left(w^{(k-T^{\left(k\right)})}\right)-F_{j^{(k)}}\left(w^{*}\right)|\,X_{0},\,X_{1},\,...,\,X_{k-T^{\left(k\right)}}\right] \nonumber \\
 & \overset{}{=}\sum_{i=1}^{N}\left(F_{i}\left(w^{(k-T^{\left(k\right)})}\right)-F_{i}\left(w^{*}\right)\right)P\left(j^{(k)}=i\,|\,X_{0},\,X_{1},\,...,\,X_{k-T^{\left(k\right)}}\right) \nonumber \\
 & {=}\sum_{i=1}^{N}\left(F_{i}\left(w^{(k-T^{\left(k\right)})}\right)-F_{i}\left(w^{*}\right)\right)P\left(j^{(k)}=i\,|\,X_{k-T^{\left(k\right)}}\right) \nonumber \\
 & \overset{}{=}\sum_{i=1}^{N}\left(F_{i}\left(w^{(k-T^{\left(k\right)})}\right)-F_{i}\left(w^{*}\right)\right)P^{(0,T^{(k)})}\left[X_{k-T^{\left(k\right)}}\,|\,j^{(k)}=i\right] \nonumber \\
 & {\geq}\left(F\left(w^{(k-T^{\left(k\right)})}\right)-F\left(w^{*}\right)\right)-\frac{N}{2k}-\frac{cte.N}{\sqrt{k}}.
 \label{eq18}
\end{align}

Finally,
\begin{align*}&\sum_{k=1}^{T}\gamma^{(k)}\mathbb{E}
\left[F\left(w^{(k)}\right)-F\left(w^{*}\right)\right]\\ & \quad \leq\sum_{k=1}^{T}N\gamma^{\left(k\right)}\left(\frac{1}{2k}+\frac{cte}{\sqrt{k}}\right)+\frac{C.D^{2}}{\ln\left(1/\lambda\right)} \\ &  \quad + \frac{1}{2}\left\Vert w^{(0)}-w^{*}\right\Vert _{2}^{2}+\sum_{k=1}^{T}{(\gamma^{(k)})}^{2}\mathbb{E}\left[\left\Vert \hat{\nabla}F_{i^{(k)}}\left(w^{(k)}\right)\right\Vert _{2}^{2}\right].\end{align*}

Using previous results and the the convexity assumption, we get 

\begin{align*} & \mathbb{E}\left[F\left(w^{(k)}\right)-F\left(w^{*}\right)\right]\\
 & \leq\frac{\sum_{k=1}^{T}N\gamma^{\left(k\right)}\left(\frac{1}{2k}+\frac{cte}{\sqrt{k}}\right)+\frac{C.D^{2}}{\ln\left(1/\lambda\right)}+\frac{1}{2}\left\Vert w^{(0)}-w^{*}\right\Vert _{2}^{2}}{\sum_{k=1}^{T}\gamma^{(k)}}\\
 & \quad+\frac{{(\gamma^{(0)})}^{2}C^{*}}{\sum_{k=1}^{T}\gamma^{(k)}}.
\end{align*}

Employing the assumptions on the step size, we get an order of convergence $O(T^{1-q})$ for a step size choice $\gamma^{(k)}=\frac{1}{k^q}$ where $\frac{1}{2}< q< 1$.


\end{document}